\def\year{2025}\relax
\documentclass[letterpaper]{article} 
\usepackage{aaai22}  
\usepackage{times}  
\usepackage{helvet}  
\usepackage{courier}  
\usepackage[hyphens]{url}  
\usepackage{graphicx} 
\usepackage{natbib}  
\usepackage{caption} 
\DeclareCaptionStyle{ruled}{labelfont=normalfont,labelsep=colon,strut=off} 
\usepackage{algorithm}
\usepackage{algorithmic}

\usepackage{xcolor}
\usepackage{booktabs}
\usepackage{amsmath}
\usepackage{amsfonts}
\usepackage{multirow}
\usepackage{array}
\usepackage{bm}
\usepackage{pbox}
\usepackage{makecell}

\usepackage{newfloat}
\usepackage{listings}

\usepackage{xcolor}
\newcommand{\answerYes}[1]{\textcolor{blue}{#1}} 
 
\newcommand{\answerNA}[1]{\textcolor{gray}{#1}}

\floatstyle{ruled}
\newfloat{listing}{tb}{lst}{}
\floatname{listing}{Listing}
\title{Demystifying Hateful Content: Leveraging Large Multimodal Models for \\ Hateful Meme Detection with Explainable Decisions}
\author {
    Ming Shan Hee,
    Roy Ka-Wei Lee
}
\affiliations {
    Singapore University of Technology and Design \\
    mingshan\_hee@mymail.sutd.edu.sg, roy\_lee@sutd.edu.sg,
}

\usepackage{bibentry}

\begin{document}

\maketitle

\begin{abstract}
Hateful meme detection presents a significant challenge as a multimodal task due to the complexity of interpreting implicit hate messages and contextual cues within memes. Previous approaches have fine-tuned pre-trained vision-language models (PT-VLMs), leveraging the knowledge they gained during pre-training and their attention mechanisms to understand meme content. However, the reliance of these models on implicit knowledge and complex attention mechanisms renders their decisions difficult to explain, which is crucial for building trust in meme classification. In this paper, we introduce \textsf{IntMeme}, a novel framework that leverages Large Multimodal Models (LMMs) for hateful meme classification with explainable decisions. \textsf{IntMeme} addresses the dual challenges of improving both accuracy and explainability in meme moderation. The framework uses LMMs to generate human-like, interpretive analyses of memes, providing deeper insights into multimodal content and context. Additionally, it uses independent encoding modules for both memes and their interpretations, which are then combined to enhance classification performance. Our approach addresses the opacity and misclassification issues associated with PT-VLMs, optimizing the use of LMMs for hateful meme detection. We demonstrate the effectiveness of \textsf{IntMeme} through comprehensive experiments across three datasets, showcasing its superiority over state-of-the-art models.
\end{abstract}


\section{Introduction}
\begin{figure}[t!]
\centering
\includegraphics[width=0.95\linewidth]{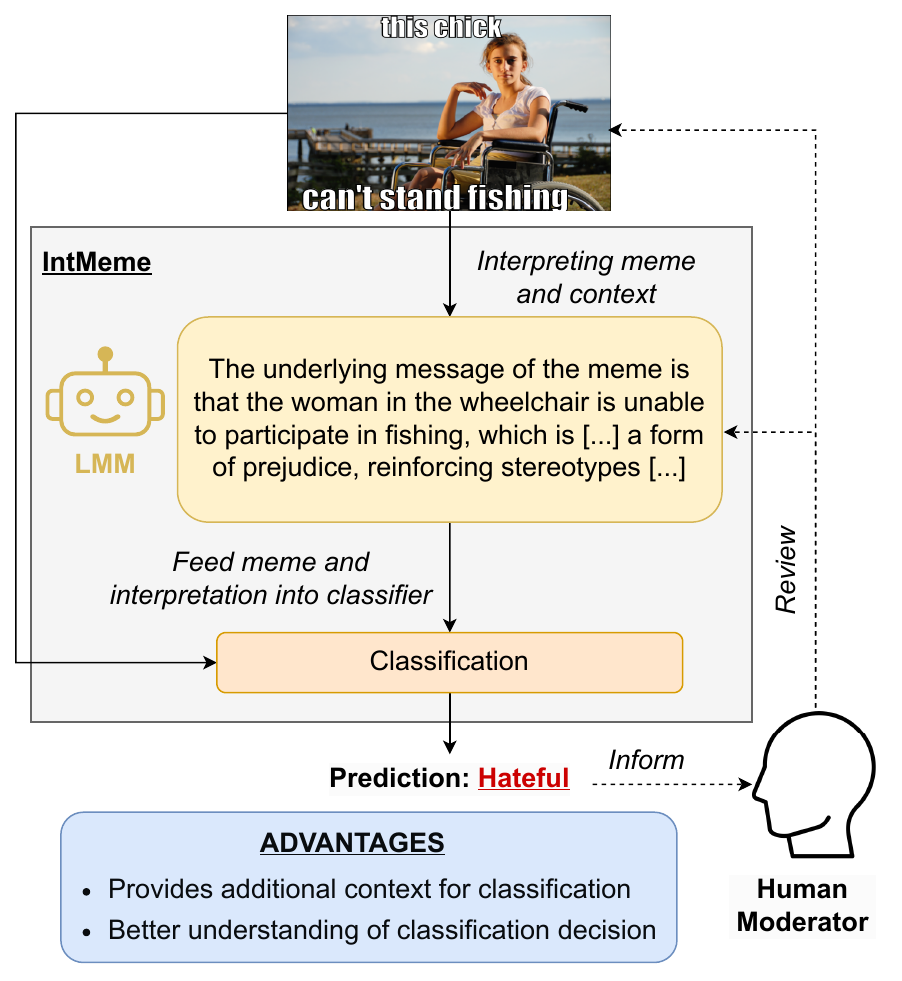}
\caption{Overview of the proposed \textsf{IntMeme}'s approach and its advantages in a content moderation process.}
\label{fig:intmeme-teaser}
\end{figure}

The rise of internet memes has significantly influenced modern communication and culture, blending humour and satire. However, the emergence of \textit{hateful} memes\footnote{\color{red}\textbf{WARNING}: \textit{This paper contains violence and discriminatory content that may be disturbing to some readers.}} reveals a darker side, contributing to social tensions, stereotyping, and misinformation. This phenomenon underscores the urgency of developing effective classification strategies to curb their negative societal impacts, highlighting the importance of promoting a harmonious and inclusive online environment.


The classification of hateful memes requires a nuanced understanding of visual and textual elements \cite{kiela2021hateful}, as well as the context behind the implied message \cite{hee2023decoding}. To address this challenge, previous research has used pre-trained vision-language transformer models (PT-VLMs) \cite{lu2019vilbert,li2019visualbert} for hateful meme classification, often enhancing these models with additional inputs like image captions \cite{velioglu2020detecting, zhu2020enhance}. While PT-VLMs can learn the interactions between visual and textual modalities, they face several limitations. Their performance heavily depends on the implicit knowledge acquired during pre-training and complex attention mechanisms, which can make it difficult to explain their decisions—an important factor for building trust in meme classification. The reliance on implicit knowledge complicates tracing the reasoning behind classifications, and the attention mechanisms make it hard to identify which features influence decisions.  Recent studies suggest that PT-VLMs might be too sensitive to subtle multimodal nuances, leading to the misclassification of non-hateful content \cite{cuo2022understanding, hee2022explaining}. These findings raise concerns about whether these models truly capture the deeper meanings that memes often convey. As a result, there is a growing need for more interpretable and efficient approaches to accurately classify hateful memes while providing clear justifications for their predictions, thereby fostering greater transparency and accountability in automated content moderation systems.

The emergence of Large Multimodal Models (LMMs), such as GPT-4(V)~\cite{openai2023gpt4v}, mPLUG-Owl~\cite{ye2023mplug}, and InstructBLIP~\cite{dai2023instructblip}, has shown promising generative capabilities. These models demonstrate strong multimodal understanding and text generation skills, which can be seen in their ability to deliver accurate and relevant text responses for complex vision-language tasks like visual question answering and visual commonsense reasoning \cite{xu2023lvlm, yang2023dawn}. Consequently, LMMs are becoming a promising solution for detecting hateful memes, providing insightful explanations into the implicit meaning hidden within memes. However, these models face several challenges. First, when used in a zero-shot setting, LMMs often perform less effectively compared to smaller models fine-tuned specifically for hateful meme classification \cite{lin2024goat}. Additionally, their zero-shot responses can sometimes deviate from the intended query, and extracting classification decisions from their generated text can be difficult, as the relevant information may be scattered throughout the output. Most importantly, fine-tuning these models demands significant computational resources and may reduce their generalizability, raising concerns about their feasibility and scalability in real-world applications. These limitations have led us to explore new methods for leveraging LMMs in hateful meme classification while maintaining their ability to produce high-quality text-based responses.


In this paper, we introduce \textsf{IntMeme}\footnote{https://github.com/Social-AI-Studio/IntMeme}, a new framework that leverages the generative abilities of large multimodal models (LMMs) to generate high-quality interpretations of memes for classifying hateful content. \textsf{IntMeme} prompts LMMs to generate these interpretations, thereby enhances the explainability of the classification process and reduces the dependence on the model's implicit knowledge. The framework then encodes both the meme and its interpretation using separate modules, which are subsequently used for final classification. This method of grounding classification decisions in meme interpretations significantly improves the accuracy and explainability of hateful meme detection while also providing clearer insights into the reasoning behind classification decisions. Figure\ref{fig:intmeme-teaser} illustrates the benefits of the \textsf{IntMeme} framework in a content moderation process involving a human moderator.

To demonstrate the effectiveness of \textsf{IntMeme} in classifying hateful memes, we conducted comprehensive experiments on three well-known datasets containing hateful memes. Our comparisons with leading PT-VLMs showed that \textsf{IntMeme} outperformed state-of-the-art baselines across all three datasets. Additionally, our ablation studies highlighted the importance of generating high-quality interpretations and utilizing distinct encoding modules, both of which significantly improved the detection of hateful memes. Furthermore, our detailed case analysis and human evaluation study underscored the effectiveness and practical utility of \textsf{IntMeme} within a simulated content moderation process. The high-quality meme interpretations used in our method enhance the explainability of the classification process, providing clearer insights into how \textsf{IntMeme} differentiates between hateful and non-hateful content. Overall, our framework promotes effective collaboration between humans and AI in online content moderation, emphasizing its contribution to the ongoing discussion about AI's role in society.

We summarize our contributions as follows: (i) We introduced \textsf{IntMeme}, a novel multimodal framework leveraging LMMs to generate insightful meme interpretations. This approach improves model explainability and provides deeper insights into the decision-making process, aiding in the distinction between hateful and non-hateful content; (ii) \textsf{IntMeme} addresses the limitations inherent in fine-tuning PT-VLMs by employing separate modules for efficient encoding of memes and their interpretations; (iii) Our comprehensive experiments on three popular harmful meme datasets validate both the efficacy and explainability of \textsf{IntMeme} when compared against similarity sized state-of-the-art harmful meme detection models. 

\section{Related Works}

\begin{figure*}[t!]
\centering
\includegraphics[width=\linewidth]{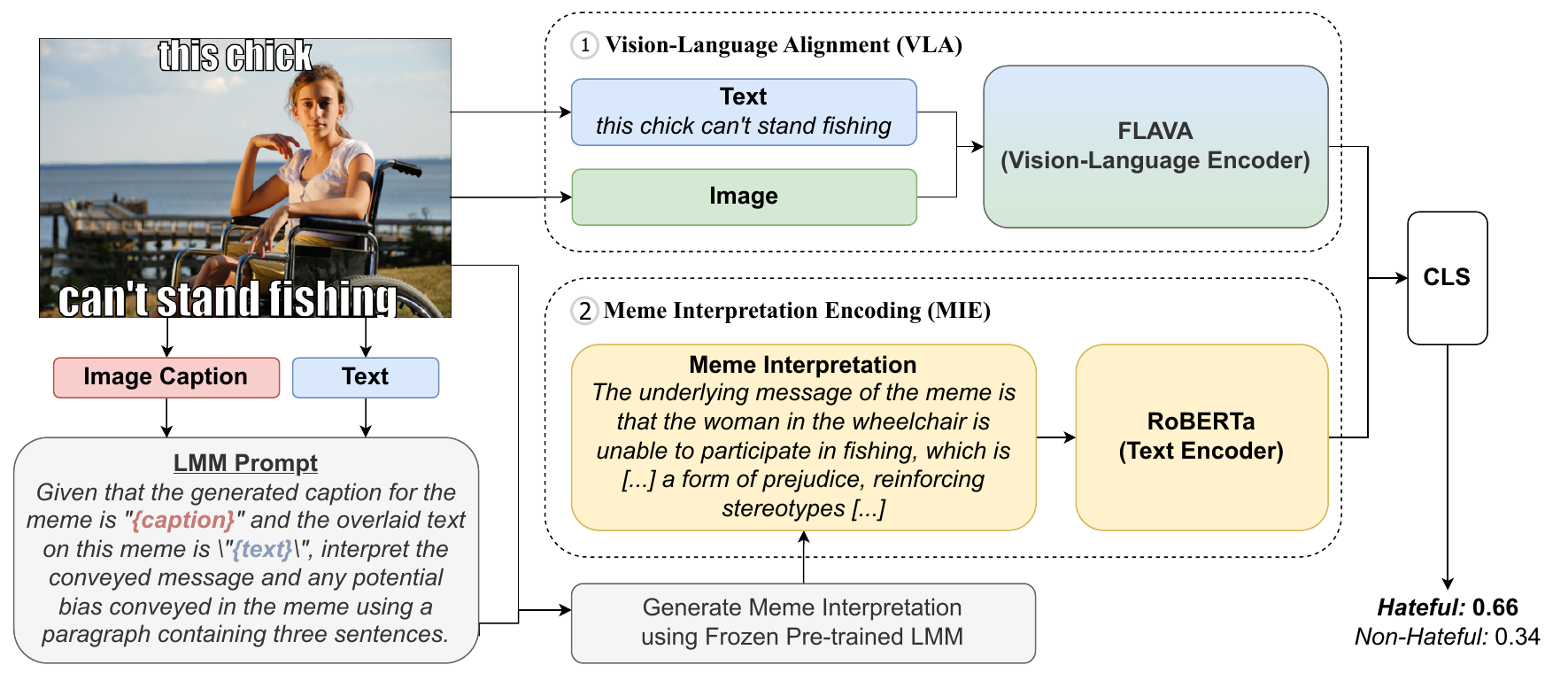}
\caption{Overview of the IntMeme framework for hateful meme classification, comprising two modules: (1) Vision-Language Alignment and (2) Meme Interpretation Encoding.}
\label{fig:framework}
\end{figure*}

\label{sec:related}
\subsection{Hate Speech Detection}
Hate speech is an increasingly prevalent issue worldwide, spreading rapidly through digital platforms and fostering social divisions. It poses a significant risk, as it not only perpetuates discriminatory attitudes but also has the potential to escalate into offline hate crimes \cite{lupu2023offline,muller2021fanning,muller2023hashtag}, resulting in severe consequences for affected communities. Researchers tackle this issue by creating datasets and developing new approaches to detect hate speech \cite{davidson2017automated,founta2018large,yoder2022hate} and explain the underlying implicit messages \cite{sap2019social,elsherief2021latent}. More recently, several studies have highlighted concerns in HS detection systems, introducing functional tests for evaluating HS detection models \cite{ng2024sghatecheck,rottger2020hatecheck,rottger2022multilingual}. Such efforts are crucial for promoting transparency and accountability, creating a safer and inclusive online environment.

Internet memes play a crucial role in online communication, serving both as sources of humor and as vehicles for disseminating hateful messages \cite{hee2024recent,hee2024understanding,uyheng2020visualizing}. This dual nature has attracted considerable attention from both industry and academia \cite{kiela2020hateful, pramanick2021harmemes, fersini2022mami, lim2024aisg, thapa2024ruhate}. The negative impacts of these hateful memes have led to the development of classification models aimed at identifying them \cite{pramanick2021momenta, thakur2022multimodal, lee2021disentangling, hee2024bridging}. For instance, \citet{pramanick2021momenta} introduced a model that detects hate by capturing complex multimodal interactions through the integration of local and global information, while also utilizing object proposals and extracted entity data. Similarly, \citet{cao2022prompting} developed a prompt-based transformer model that incorporates examples of both hateful and non-hateful memes, along with an unseen meme for inference. More recent studies have employed large multimodal models (LMMs) to generate explanations based on the true classification labels in the prompt, followed by fine-tuning a classification model using knowledge distillation \cite{lin2023beneath} or multimodal debate mechanisms \cite{lin2024towards}. In contrast, IntMeme prompts LMMs to produce meaningful interpretations of memes without relying on knowledge of the true labels and introduces separate modules to efficiently encode both the meme and its interpretation.

\subsection{Large Multimodal Models}
The emergence of large multimodal models (LMMs) represents a significant advancement in artificial intelligence, demonstrating impressive generative capabilities \cite{ye2023mplug,dai2023instructblip,openai2023gpt4v,deitke2024molmo}. These models typically utilize a pre-trained large language model (LLM) as their foundational base, incorporating a vision projection module that includes an image encoder (e.g., ViT \cite{dosovitskiy2020image}, EVA \cite{fang2023eva}) and several image projection layers to convert images into the text embedding space. This configuration enables LMMs to effectively interpret visual inputs while leveraging the robust language modeling capabilities of the foundational LLM.

LMMs have demonstrated excellent performance across various multimodal tasks, including understanding the subtleties of humor within visual and textual content \cite{yang2023dawn,zheng2023judging,han2023evaluate}. Recent research on LMMs has also explored reasoning capabilities related to humor, focusing specifically on jokes and humorous memes. For instance, \citet{ye2023mplug} demonstrate that their mPLUG-Owl LMM exhibits strong vision-language understanding, allowing it to grasp visually driven jokes. Additionally, \citet{yang2023dawn} highlight the impressive ability of GPT-4V to extract information from both visual and textual modalities, facilitating the comprehension of humor within memes. Drawing inspiration from Socratic models \cite{zeng2022socratic} that combine various large pre-trained models to address new multimodal challenges, our approach employs LMMs to generate meaningful interpretations prior to their use in classifying hateful memes. This strategy enhances both model performance and explainability.


\section{Methodology}

The \textsf{IntMeme} framework uses the strong multimodal reasoning capabilities of LMMs to generate high-quality interpretations of memes, aiding in the classification of hateful memes. This approach mirrors the human process of understanding memes before assessing their potential for hatefulness. Additionally, this approach enables end-users to review and comprehend the generated interpretations, enhancing the explainability of the classification process. Figure \ref{fig:framework} presents an overview of the \textsf{IntMeme} framework.


\subsection{Generating Meme Interpretation}

\paragraph{Zero-Shot Inference using LMMs.} Our methodology uses a pre-trained language model in a zero-shot setting. Internet memes, created and shared by diverse netizens, cover numerous topics, tones, and cultural contexts. \cite{kiela2020hateful,fersini2022mami}. This variety makes it challenging to adequately address all the variations with a limited set of demonstration examples. A limited set of examples may introduce bias, hindering the model's ability to interpret diverse content accurately and increasing computational resources and processing time. On the other hand, recent studies also highlighted the strong zero-shot reasoning capabilities of LMMs across various multimodal tasks \cite{yang2023dawn,fu2023mme}. Hence, our approach leverages and explores the limits of LMMs in a zero-shot setting.


\definecolor{text_placeholder}{HTML}{8A9DB8}
\definecolor{length_control_measure}{HTML}{1B9E77}
\definecolor{caption_placeholder}{HTML}{C77975}

\begin{table}[t]
\small
\centering
\begin{tabular}{p{\linewidth}}
\toprule
\textbf{System Instructions*} \\
\midrule
The following is a conversation between a human content moderator, who works on meme moderation, and an AI assistant. The assistant provides an informative interpretation of memes, including details about the underlying message and any potential prejudice (i.e. towards individuals or communities) within the memes. It is important that the interpretation utilizes both the visual and linguistic elements of the memes. \\ 
\bottomrule \\
\toprule
\textbf{Human Prompt} \\ 
\midrule
Given that the generated caption for the meme is ``\textcolor{caption_placeholder}{\{caption\}}'' and the overlaid text on this meme is ``\textcolor{text_placeholder}{\{text\}}'', interpret the conveyed message and any potential bias conveyed in the meme \textcolor{length_control_measure}{using a paragraph containing three sentences}. \\
\bottomrule
\end{tabular}
\caption{Example system instructions and prompts for generating \textsl{meme interpretation}. The prompt input includes the caption placeholder tag (in \textcolor{caption_placeholder}{orange}), text placeholder tag (in \textcolor{text_placeholder}{blue}), and the length control measure (in \textcolor{length_control_measure}{green}). \textit{*InstructBLIP does not customization of model behaviour via system instructions.}}
\label{tab:lvlm_settings}
\end{table}

\paragraph{System Instructions.} 
To ensure the generation of accurate and high-quality meme interpretations using a zero-shot approach, we implemented a careful process that involved customizing the behavior of a large multimodal model (LMM) and creating a meaningful human prompt. By employing custom system instructions, we were able to adjust the responses of these instruction-tuned models, aligning them with the objectives of meme interpretability. These instructions guide the models to produce informative interpretations while effectively identifying potential visual and textual nuances within the memes that may reflect social prejudice. The system instruction can be found in Table~\ref{tab:lvlm_settings}.

\paragraph{Prompt Design.} 
We designed a model prompt to guide the LMMs in producing clear and informative meme interpretations. The objetive of this prompt is to generate a high-quality interpretation that not only captures the meme's underlying message but also identifies any potential bias it may convey. To achieve this, we designed the prompt to include both the text overlay from the meme and an image caption generated by the same LMM, encouraging the model to focus on reasoning. However, recent studies also show that instruction-tuned LMMs often produce lengthy responses, which can lead to performance and encoding challenges, particularly with pre-trained encoding modules. To address this, we incorporated explicit length control measures into the model prompt. The details of the human prompt are provided in Table \ref{tab:lvlm_settings}.

\subsection{Information Encoding \& Classification}
The IntMeme framework uses two distinct modules to encode information efficiently: the \textsl{meme interpretation encoding} (MIE) module and the \textsl{vision-language alignment} (VLA) module. The MIE module is responsible for learning the semantic meanings of meme interpretations, whereas the VLA module focuses on capturing both the inter- and intra-modality information present within memes. Subsequently, these encoded representations are combined to classify potentially hateful memes. This approach improves the model's explainability by providing insights into the model's decisions through the conditioned meme interpretation.

\subsubsection{Meme Interpretation Encoding Module}
We use a separate text encoder module to learn the semantic meaning of the meme interpretation. Formally, we feed the generated meme interpretation $\mathcal{G}$ into the text encoder model to generate the hidden states $\mathbf{I}$:

\begin{equation*}
    \mathbf{I} = \text{Encoder}_\text{text}(\mathcal{G})
\end{equation*}

From the hidden states $\mathbf{I}$, we use the hidden state from the $\mathbf{[CLS]}$ token ($\mathbf{I}_{CLS}$). This token has demonstrated effectiveness in sentence understanding tasks, as evidenced in \cite{reimers2019sentence,liu2019roberta}.

\subsubsection{Vision Language Alignment Module.} 
While generating meme interpretations in a zero-shot approach offers practical advantages, these interpretations can be misleading or contain inaccuracies \cite{ji2023survey}. To reduce the severity of misleading interpretation, the vision-language alignment module processes and supplements the intricate inter- and intra-modality interactions within the meme. This supplementary meme information allows the model to rely on the vision and language information within the meme, alleviating the model’s dependency on the generated meme interpretation. Formally, we feed the the meme image $\mathcal{V}$ and meme text $\mathcal{T}$ into a vision-language model to generate the hidden states $\mathbf{M}$:

\begin{equation*}
    \mathbf{M} = \text{Encoder}_\text{vision-language}([\mathcal{V}, \mathcal{T}])
\end{equation*}

From the hidden states $\mathbf{M}$, we use the hidden state from the $\mathbf{[CLS]}$ token ($\mathbf{M}_{CLS}$). This token has been included to facilitate the multimodal understanding tasks during pre-training, serving as an ideal representation of the meme context.





\subsection{Classification Layer}






After obtaining the representations ($\mathbf{M}_{CLS}$ and $\mathbf{I}_{CLS}$), we concatenate and feed them into a classification layer. The classification layer consists of a single-layer perception followed by a softmax layer for normalization.

\begin{equation}
    \nonumber
    O = \text{Sigmoid}(W^\text{T}[\mathbf{M}_{CLS}, \mathbf{I}_{CLS}]+ b),
\end{equation}

where $[\cdot, \cdot]$ represents concatenation, $W \in \mathbb{R}^{d \times 2}$ are learnable weights and $b \in \mathbb{R}^{2}$ are learnable bias. The final prediction, $O \in \mathbb{R}^{2}$, represents the logits for each class.

\section{Experiment Settings}

\begin{table}[t]
\centering
  \begin{tabular}{c|cc|cc}
    \hline
     & \multicolumn{2}{c|}{\textbf{Train}} & \multicolumn{2}{c}{\textbf{Test}}\\
    Dataset & \# H & \# Non-H & \# H & \# Non-H\\
    \hline\hline
    FHM-FG & 3,007 & 5,493 & 246 & 254 \\
    HarMeme & 1,064 & 1,949 & 124 & 230\\
    MAMI & 5,004 & 4,996 & 500 & 500 \\
    \hline
\end{tabular}
\caption{Statistical distributions of datasets, where "H" represents harmful and "Non-H" represents non-harmful }
  \label{tab:dataset}
\end{table} 

\subsection{Evaluation Datasets} 
We evaluated \textsf{IntMeme} against the state-of-the-art PT-VLMs across three widely-used hateful meme datasets, showcasing its robustness and generalizability. 
\textit{Facebook’s Fine-Grained Hateful Memes} (\textbf{FHM-FG}) dataset \cite{mathias2021fhmfg} is a synthetic memes dataset containing hateful memes with five distinct types of incitement to hatred: gender, racial, religious, nationality and disability-based. \textit{Multimedia Automatic Misogyny Identification} (\textbf{MAMI}) dataset \cite{fersini2022mami} consists of misogynous memes collected from popular social media platforms and websites dedicated to meme creation. Evaluating our models on this dataset provides insight into the performance of hateful meme detection models in a natural environment. \textit{Harmful Meme} (\textbf{HarMeme}) dataset \cite{pramanick2021harmemes} consists of crowdsourced memes primarily collected from Google Image Search and publicly available groups on popular social media websites. These memes contains \textit{harmless}, \textit{partially harmful}, and \textit{very harmful} memes related to the COVID-19 topic. Following \citeauthor{pramanick2021harmemes}, we merge \textit{partially harmful}, and \textit{very harmful} into a single \textit{harmful} category.
A summary of the distribution of the three datasets is presented in Table \ref{tab:dataset}.

\subsection{Models}
We evaluated \textsf{IntMeme} against seven state-of-the-art models. The \textbf{VisualBERT} \cite{li2019visualbert} model uses a single-stream transformer-based approach that concurrently processes textual and visual inputs using a single Transformer module. In contrast, the \textbf{ViLBERT} \cite{lu2019vilbert} uses a dual-stream transformer-based approach that independently processes textual and visual inputs before using Transformer modules to capture inter-modality interactions. More recently, the \textbf{BLIP} \cite{li2022blip} model is pre-trained on a mixture of multimodal encoder-decoder models using a dataset bootstrapped from large-scale noisy image-text pairs. The \textbf{FLAVA} \cite{singh2022flava} model is pre-trained on multimodal and unimodal data with unpaired images and text. Moving into models designed for hateful memes detection, the \textbf{MOMENTA} \cite{pramanick2021momenta} model utilizes both local and global multimodal fusion mechanisms to exploit interactions for detecting harmful memes. The \textbf{DisMultiHate} \cite{lee2021disentangling} model adopts a disentanglement approach to separate target information from memes, crucial for identifying hateful content. Lastly, the \textbf{PromptHate} \cite{cao2022prompting} model uses a prompt-based approach with few-shot demonstrations to classify memes.

\subsection{Evaluation Metrics}

We employed two widely adopted metrics to evaluate the performance of the various models: Accuracy (Acc.) and Area Under the Receiver Operating Characteristics curve (AUROC). All the experimental results are aggregated across five random seeds, with the average results and standard deviation reported. All the models use the same set of random seeds to ensure a fair comparison.

\subsection{Implementation Details}
\paragraph{Large Multimodal Models.} 

We compare two open-source LMMs with robust multimodal reasoning capabilities: mPLUG-Owl \cite{ye2023mplug} and InstructBLIP \cite{dai2023instructblip}. These LMMs have shown impressive overall visual perception and cognition abilities, as evidenced by their high rankings on the MME benchmark leaderboards \cite{fu2023mme}. We prompt the pre-trained LMMs to generate the image captions before prompting them to generate the meme interpretation. For reproducibility, we use greedy decoding. Moreover, to minimize the occurrence of lengthy and repetitive responses, we configure the decoding settings to use no\_repeat\_ngram\_size = 2 and max\_new\_tokens = 256.

\paragraph{IntMeme Encoders.} 
The MIE module uses RoBERTa as its text encoder, while the VLA module employs FLAVA as the vision-language encoder. The RoBERTa model has shown proficiency across various language modelling tasks. The FLAVA model, trained on the hateful meme detection task during pre-training, is well-suited for modelling the complex inter- and intra-modality interactions within memes.

\paragraph{IntMeme Training.} 
We use a learning rate of 2e-5 and a batch size of 32 to fine-tune \textsf{IntMeme} on 1 A100 GPU over 30 epochs with early stopping (i.e., patience = 5)\footnote{The model typically converges within 10 epochs}. As for the selection of the models, we base our choices on the average of their Acc. and AUROC scores. We optimized these models using Adam optimizer \cite{kingma2015adam} and are implemented in PyTorch using the Huggingface's \texttt{Transformers}\footnote{https://huggingface.co/docs/transformers} library.

\begin{table*}[t!]
    \centering
    \begin{tabular}{ccccccc}
        \toprule
         &\multicolumn{2}{c}{\textbf{FHM}}&\multicolumn{2}{c}{\textbf{MAMI}}&\multicolumn{2}{c}{\textbf{HarMeme}}\\
         \cmidrule(lr){2-3} \cmidrule(lr){4-5} \cmidrule(lr){6-7}
        \textbf{Model} & \textbf{AUROC} & \textbf{Acc.}& \textbf{AUROC} & \textbf{Acc.} & \textbf{AUROC} & \textbf{Acc.}\\
        \midrule
        VisualBERT & 68.71$_{\pm1.02}$& 61.48$_{\pm1.19}$  &78.71$_{\pm0.59}$ &71.06$_{\pm0.94}$  &80.46$_{\pm1.04}$ &75.31$_{\pm1.44}$ \\
        ViLBERT & 73.05$_{\pm0.62}$&64.70$_{\pm1.12}$  &77.71$_{\pm1.20}$ &69.48$_{\pm1.00}$  &84.11$_{\pm0.88}$ &78.70$_{\pm1.17}$  \\
        MOMENTA$^*$ & 69.17$_{\pm4.71}$ & 61.34$_{\pm4.89}$  &81.68$_{\pm2.80}$ &72.10$_{\pm2.90}$   & 86.32$_{\pm3.83}$& 80.48$_{\pm1.95}$\\
        DisMultiHate & 69.11$_{\pm0.84}$& 62.42$_{\pm0.72}$ &78.21$_{\pm0.61}$ & 70.58$_{\pm1.13}$ & 83.69$_{\pm1.33}$& 78.05$_{\pm0.73}$ \\
        PromptHate & 76.76$_{\pm0.95}$&67.82$_{\pm1.23}$  &76.21$_{\pm1.05}$ &68.08$_{\pm0.58}$  &87.51$_{\pm0.74}$ & 79.38$_{\pm1.72}$ \\
        BLIP & 76.80$_{\pm2.37}$ &69.20$_{\pm1.84}$  & 80.59$_{\pm0.87}$&71.84$_{\pm1.11}$  &87.09$_{\pm1.46}$ &81.81$_{\pm1.74}$  \\
        FLAVA & 78.51$_{\pm0.70}$ & 70.28$_{\pm1.03}$ & 80.69$_{\pm0.84}$ & 71.72$_{\pm0.36}$ & 88.34$_{\pm1.15}$ & 81.58$_{\pm1.40}$ \\
        \midrule
        IntMeme$_\text{InstructBLIP}$ & 81.05$_{\pm0.81}$ & \textbf{71.48$_{\pm1.71}$} & 81.59$_{\pm0.65}$ & \textbf{72.44$_{\pm0.88}$} & 88.00$_{\pm0.84}$ & \textbf{82.66$_{\pm1.33}$} \\
        IntMeme$_\text{mPLUG-Owl}$ & \textbf{81.50$_{\pm1.11}$} & 71.52$_{\pm1.49}$ & \textbf{81.89$_{\pm1.15}$} & 72.30$_{\pm1.79}$ & \textbf{89.35}$_{\pm1.22}$ & 81.92$_{\pm2.47}$ \\
        \bottomrule
    \end{tabular}
    \caption{Evaluation results of hateful meme detection models on three benchmark datasets \textbf{without} any augmented image tags. These results have been aggregated over 5 random seeds and are reported along with their corresponding standard deviations.}
    \label{tab:experimental-results}
\end{table*}

\begin{table*}[t]
  \centering
  \begin{tabular}{lcccccc}
    \toprule
     &\multicolumn{2}{c}{\textbf{FHM}}&\multicolumn{2}{c}{\textbf{MAMI}}&\multicolumn{2}{c}{\textbf{HarMeme}} \\
     \cmidrule(lr){2-3} \cmidrule(lr){4-5} \cmidrule(lr){6-7}
    \textbf{Model} &\textbf{AUC.}&\textbf{Acc.} &\textbf{AUC.}&\textbf{Acc.} &\textbf{AUC.}&\textbf{Acc.} \\
    \midrule
    IntMeme$_\text{InstructBLIP}$ & & \\
    $-$  w/ \textsc{INTPN (MIE Module)} & 75.49$_{\pm1.46}$ & 68.64$_{\pm1.56}$ & 75.22$_{\pm1.56}$ & 66.50$_{\pm2.26}$ & 83.04$_{\pm1.96}$ & 77.12$_{\pm2.14}$  \\
    $-$  w/ \textsc{Meme (VLA Module)} & 78.51$_{\pm0.70}$ & 70.28$_{\pm1.03}$ & 80.69$_{\pm0.84}$ & 71.72$_{\pm0.36}$ & \textbf{88.34$_{\pm1.15}$} & 81.58$_{\pm1.40}$ \\
    $-$  w/ \textsc{Both (MIE + VLA Module)} & \textbf{81.05$_{\pm0.81}$} & \textbf{71.48$_{\pm1.71}$} & \textbf{81.59$_{\pm0.65}$} & \textbf{72.44$_{\pm0.88}$} & 88.00$_{\pm0.84}$ & \textbf{82.66$_{\pm1.33}$} \\
    \midrule
    IntMeme$_\text{mPLUG-Owl}$ & & \\
    $-$  w/ \textsc{INTPN (MIE Module)} & 77.26$_{\pm0.66}$ & 68.24$_{\pm2.42}$ & 77.61$_{\pm0.91}$ & 70.18$_{\pm0.72}$ & 88.74$_{\pm1.77}$ & 78.81$_{\pm2.32}$  \\
    $-$  w/ \textsc{Meme (VLA Module)} & 78.51$_{\pm0.70}$ & 70.28$_{\pm1.03}$ & 80.69$_{\pm0.84}$ & 71.72$_{\pm0.36}$ & 88.34$_{\pm1.15}$ & 81.58$_{\pm1.40}$ \\
    $-$  w/ \textsc{Both (MIE + VLA Module)} & \textbf{81.50$_{\pm1.11}$} & \textbf{71.52$_{\pm1.49}$} & \textbf{81.89$_{\pm1.15}$} & \textbf{72.30$_{\pm1.79}$} & \textbf{89.35$_{\pm1.22}$} & \textbf{81.92$_{\pm2.47}$} \\
    \midrule
    FLAVA \\
    $-$ \textsc{ Vanilla} & 78.51$_{\pm0.70}$ & 70.28$_{\pm1.03}$ & 80.69$_{\pm0.84}$ & 71.72$_{\pm0.36}$ & 88.34$_{\pm1.15}$ & 81.58$_{\pm1.40}$ \\
    $-$ w/ \textsc{INTPN}$_\text{InstructBLIP}$ \textsc{(CONCAT)} & 78.98$_{\pm0.79}$ & \textbf{70.52}$_{\pm0.87}$ & \textbf{81.23}$_{\pm1.28}$ & \textbf{71.22}$_{\pm2.59}$ & 88.63$_{\pm0.78}$ & 80.73$_{\pm2.79}$ \\
    $-$ w/ \textsc{INTPN}$_\text{mPLUG-Owl}$ \textsc{(CONCAT)} & \textbf{79.45}$_{\pm0.85}$ & 70.44$_{\pm1.58}$ & 81.20$_{\pm1.03}$  & 70.84$_{\pm2.22}$ & \textbf{89.10}$_{\pm1.16}$ & \textbf{81.53}$_{\pm2.32}$  \\
    \bottomrule
\end{tabular}
\caption{Ablation study w.r.t \textsf{IntMeme} and its distinct modules. The top scores across the variations are highlighted in \textbf{bold}.}
\label{tab:ablation-modules}
\end{table*}

\section{Experiments}
\subsection{Hateful Meme Classification} 

Table \ref{tab:experimental-results} displays the evaluation results of state-of-the-art baselines on three benchmark datasets. We report the average score and standard deviation across five random seeds and highlight the best performance in bold. Both the IntMeme$_\text{InstructBLIP}$ and IntMeme$_\text{mPLUG-Owl}$ variants outperform the state-of-the-art baselines across all three datasets, improving by 2.54, 0.9, and 1.01 percentage points in absolute AUC performance, respectively. The superior performance and low standard deviation underscore the effectiveness of our proposed framework in the hateful meme detection task. Notably, both model variants consistently achieve better performance, with IntMeme$_\text{InstructBLIP}$ securing the highest accuracy and IntMeme$_\text{mPLUG-Owl}$ the best AUC performance across the datasets. These results suggest that both the mPLUG-Owl and InstructBLIP LMMs excel in generating highly informative meme interpretations that enhance hateful meme detection. Nevertheless, the informativeness and effectiveness of these interpretations warrant further analysis, which we will discuss in the empirical analysis section.

\subsection{Ablation Study}


We conducted two ablation studies to examine the effectiveness of generated meme interpretations and distinct encoding modules. In the first study, we evaluated the effectiveness of the generated interpretations by comparing the performance of three setups: using only the generated interpretations (the ``MIE module''), using only the meme (the ``VLA module''), and using both the interpretations and the meme together (the "MIE + VLA modules"). The second study focused on the importance of separate encoding modules, comparing a fine-tuned FLAVA model, which uses concatenated meme and interpretation data for hateful meme classification, against the \textsf{IntMeme} model. Table \ref{tab:ablation-modules} presents the results of these ablation studies for both the \textsf{IntMeme} and FLAVA models.

\paragraph{Meme Interpretation.} 

Firstly, we observe that, despite a notable decrease in performance, the \textsf{IntMeme} model variant fine-tuned solely with meme interpretations achieves performance levels comparable to the state-of-the-art PT-VLMs baselines. This highlights the informative nature of the generated meme interpretations, suggesting their potential as competitive substitutes for visual cues and modality interactions in memes. Our investigation further reveals that both model variants, which combine meme interpretations with meme data, significantly outperform their counterparts trained solely on memes (with improvements of 2.54 and 2.99 percentage points in absolute AUROC performance, respectively). Additionally, the FLAVA model fine-tuned with meme interpretations from InstructBLIP or mPLUG-Owl outperforms the variant fine-tuned solely on memes, with gains of 0.47 and 0.94 percentage points in absolute AUROC performance, respectively. These findings underscore the informative value of the generated meme interpretations.

\begin{table*}[t]
  \small
  \centering
    \resizebox{1\linewidth}{!}{$
  \begin{tabular}{c|p{8.1cm}|p{8.1cm}}
  \hline
  
    \multirow{2}{*}{\textbf{Memes}}
    
    & \begin{minipage}[!b]{0.80\columnwidth}
      \centering
      \raisebox{-.7\height}{\includegraphics[width=\linewidth]{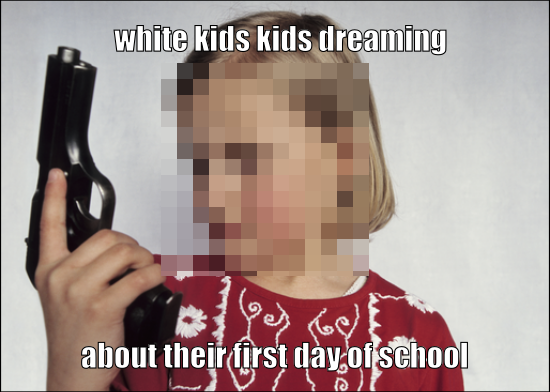}}
         \newline (a)
     \end{minipage}  
     &
    \begin{minipage}[!b]{0.80\columnwidth}
  \centering
  \raisebox{-.5\height}{\includegraphics[width=\linewidth]{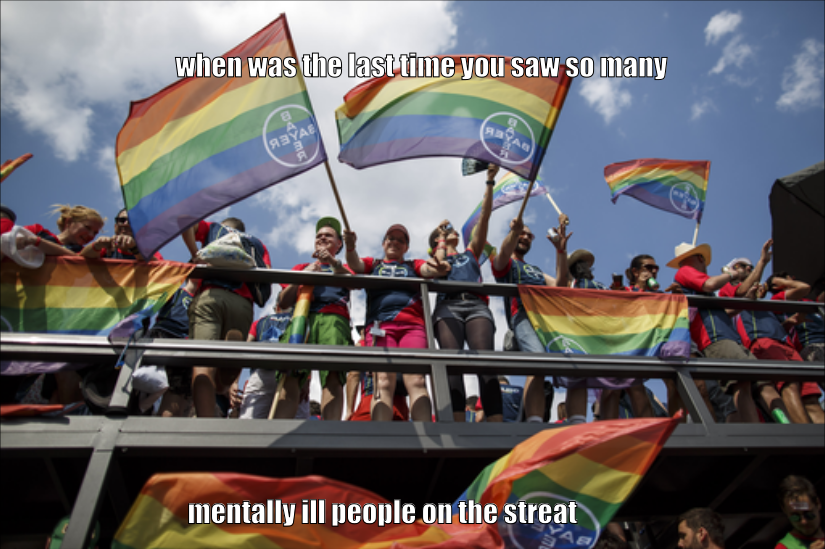}}
         \newline (b)
 \end{minipage}\\
    \hline 

    \textbf{\makecell{Interpretation \\ (mPLUG-Owl)}} &
    The underlying message of the meme is that the young girl is holding a gun and posing for a picture, possibly during their first day of school. This could be interpreted as a playful or rebellious act, as guns are not typically associated with school settings. Additionally, the overlaid text ``white kids dreaming" implies that white children are more likely to engage in such activities or have a different perception of guns compared to children from other racial or ethnic backgrounds. The potential bias conveyed in the image is the stereotypical association of white people with guns and violence, reinforcing the idea that they are perceived as more aggressive or prone to dangerous behavior. &
    The underlying message of the meme seems to be that the presence of a large crowd of people holding rainbow flags on the street suggests that there is a high prevalence of mental illness in the community. The overlaid text, ``when was the last time you saw so many mentally ill people on a street," implies that this is an uncommon sight and that it may be a cause for concern. However, it is important to note that mental health issues can affect people from all walks of life, and the use of stereotypes and stigmatizing language can reinforce negative perceptions and discrimination against individuals with mental disorders.
    \\
    \hline
    \end{tabular}
    $}
  \caption{Case study analysis on two randomly chosen hateful memes and their interpretations.}
  \label{tab:case-study-analysis}
\end{table*}

\begin{table}[t]
  \centering
  \begin{tabular}{lrrrrr}
    \toprule
    & \textbf{Cl.}&\textbf{Acc.} &\textbf{Rel.}&\textbf{Help.} & \textbf{Rec.} \\
    \midrule
    \#. Avg Score $\leq$ 3 & 38 & 66 & 63 & 71 & 67 \\
    \#. Avg Score $\geq$ 4 & 112 & 84 & 87 & 79 & 83 \\
    \midrule
    Mean & 3.78 & 3.21 & 3.38 & 3.15 & 3.23 \\
    Median & 4.00 & 3.33 & 3.67 & 3.33 & 3.33 \\
    Mode & 4.67 & 4.00 & 5.00 & 5.00 & 4.67 \\
    \bottomrule
\end{tabular}
\caption{Statistic breakdown of the human evaluation results for 150 memes, evenly sampled from FHM, HarMeme, and MAMI datasets: Clarity (Cl.), Accuracy (Acc.), Cultural Relevance (Rel.), Helpfulness (Help.) and Recognition of Hateful Elements (Rec.).}
\label{tab:human-evaluation}
\end{table}

\paragraph{Encoder Modules.} 
The superior performance of both \textsf{IntMeme} variants (IntMeme$\text{InstructBLIP}$ and IntMeme$\text{mPLUG-Owl}$) compared to the FLAVA model fine-tuned for meme interpretation, with absolute score improvements of 2.05 and 2.07, highlights the benefits of using separate encoder modules for processing the meme and its interpretation. Importantly, removing either the Visual Language Adapter (VLA) or the Meme Interpretation Encoder (MIE) from IntMeme's architecture leads to a noticeable drop in performance, particularly when the VLA module is omitted. The VLA module likely addresses the inaccuracies or gaps in meme interpretation and enhances the model's ability to integrate information across modalities. This finding emphasizes the importance of incorporating a VLA module for effectively encoding meme interactions, further supporting its role in the task of hateful meme classification.

\section{Empirical Analysis}


We conducted a human evaluation study and performed a case study analysis to understand the real-world utility of meme interpretation in explaining IntMeme's decisions. For the case study analysis, we randomly sampled two hateful memes misclassified by FLAVA but correctly classified by IntMeme. Subsequently, we used LIME \cite{lime}, a model-agnostic explainer, to understand and explain the influence of meme interpretations on the IntMeme$_\text{mPLUG-Owl}$'s decisions. Lastly, we will discuss the limitations and future directions for this line of work involving generative LMMs.

\subsection{Human Evaluation Study} 

\paragraph{Study Design.} 

This study aims to evaluate the quality and utility of meme interpretations based on \textit{clarity}, \textit{accuracy}, \textit{cultural relevance}, and their \textit{helpfulness} in identifying hateful content, employing a 5-point Likert scale for assessment. Quality metrics focus on the ease of understanding, faithfulness to the meme's intended message, and alignment with cultural context, while utility is measured by the interpretations' helpfulness in revealing the meme's message and detecting hatefulness. The evaluation was conducted by three English-proficient university students, ensuring a rigorous examination of the generated interpretations' impact on understanding and classifying memes.


\paragraph{Study Execution.}

The human evaluators assessed 150 hateful memes sourced equally from the FHM, HarMeme, and MAMI datasets. To standardize their evaluations, the evaluators participated in two preliminary rounds of review, aimed at harmonizing their assessment criteria. To further ensure the study's reliability, we introduced 15 control questions featuring memes with deliberately mismatched interpretations. Evaluators are expected to recognize these incongruities and assign low-quality scores, thereby validating the consistency and reliability of their assessments.


\paragraph{Results Analysis.}

Table \ref{tab:human-evaluation} details the results of our human evaluation study, summarized by average scores assigned to each meme interpretation on a 5-point scale. The interpretations of most hateful memes scored above 3, demonstrating their effectiveness and utility. Furthermore, the analysis of central tendency measures indicates a left-skewed distribution across all evaluated metrics. This skewness implies that a majority of the interpretations were rated highly, receiving scores on the upper end of the scale, with fewer instances of low scores. Such a distribution underscores the interpretations' success in accurately conveying the memes' intended messages, affirming their overall effectiveness.


\subsection{Case Study Analysis}

\paragraph{Meme Interpretation.} 

Table \ref{tab:case-study-analysis} showcases the details of randomly chosen hateful memes and their corresponding interpretations from the FHM and MAMI datasets, respectively. We notice that the generated interpretation of meme \ref{tab:case-study-analysis}(a) effectively utilizes both textual and visual information to depict the meme. Subsequently, the interpretation captures the underlying hate implication that white children are more prone to acts of violence or terrorism, stemming from a stereotypical bias associated with white individuals. On the other hand, the interpretation for meme \ref{tab:case-study-analysis}(b) contains a high level of inaccuracies. Although the interpretation initially connects the idea of a ``large crowd of people waving rainbow flags'' with a ``high prevalence of mental illness in the community'', it mistakenly assumes that the meme discusses mentally ill people on the street. This misunderstanding completely alters the implicit hate message, distorting the original intention of the meme. Nevertheless, the generated interpretation still discourages the use of stereotypical bias and stigmatizing language against people with mental disorders.

\begin{figure}[t]
\centering
\includegraphics[width=\linewidth]{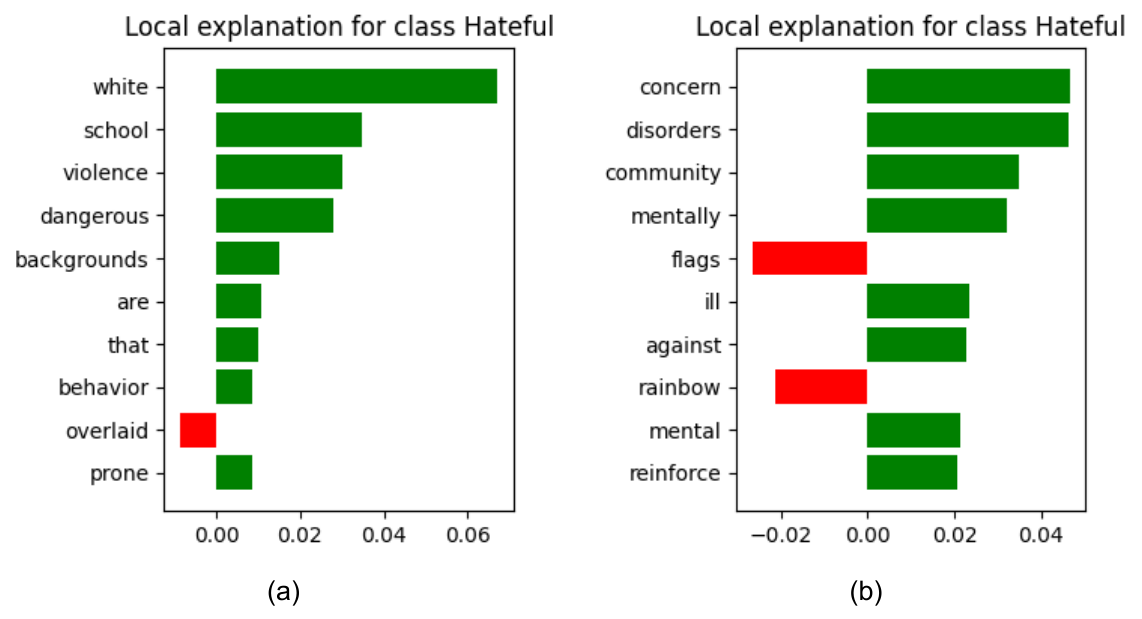}
\caption{LIME's visualization of the meme interpretation’s contribution towards IntMeme$_\text{mPLUG-Owl}$ model's prediction}
\label{fig:lime-visualization}
\end{figure}

\paragraph{Enhanced Model Explainability.} 

IntMeme$_\text{mPLUG-Owl}$ conditions the classification of hateful memes based on both the meme and its interpretation. Therefore, we utilized \textit{Locally Interpretable Model-Agnostic Explanations} \cite{ribeiro2016should}, a model-agnostic explainer, to further explain the influence of meme interpretations on the model’s classification results. The visualization of the interpretation’s contribution to the IntMeme$_\text{mPLUG-Owl}$ model’s classification is illustrated in Figure \ref{fig:lime-visualization}. We observed that stereotypical related terms such as “white”, “dangerous”, “school” and “violence” contributes significantly to the model’s classification for meme \ref{tab:case-study-analysis}(a), which aligns to our case study findings. Similarly, words related to mental disability, such as "mentally," "mental," and "ill," play a substantial role in the model's classification for meme \ref{tab:case-study-analysis}(b). It is important to highlight that many of these highly contributing words are absent in the original meme's text, underscoring how the generated interpretations assist in clarifying the model's decision, which would otherwise be challenging to explain. In summary, our LIME analysis reinforces our belief that having meme interpretation is useful for improving and explaining the classification of hateful meme.

\section{Discussion and Conclusion}

The deployment of LMMs for multimodal downstream tasks is fraught with challenges: the fine-tuning process is resource-intensive, deriving specific answers from generated texts can be cumbersome, and the models are prone to hallucinating content \cite{ji2023survey}. In response, this study proposes \textsf{IntMeme}, an innovative framework designed to efficiently utilize LMMs for generating insightful interpretations of memes, particularly to aid in classifying hateful content.

Our empirical investigation, which includes a human evaluation study and a manual examination, reveals that while LMMs can produce quality interpretations, there remains a considerable margin for improvement. Specifically, we identified three recurrent issues: inaccuracies in visual element identification leading to confusion, failures in detecting sarcasm or wordplay resulting in overly literal interpretations, and the issue of incomplete interpretations due to premature model termination. These findings underscore the limitations of open-source LMMs in meme interpretation and open pathways for further research advancement.

In conclusion, \textsf{IntMeme} presents a novel framework that uses LMM to improve the performance and explainability of hateful meme classification. \textsf{IntMeme} prompts LMM in a zero-shot manner before using separate modules to encode the meme and the LMM-generated interpretation efficiently. Our comprehensive experiments on three popular harmful meme datasets demonstrate the framework's effectiveness. Despite the high quality and utility of most generated meme interpretations, our study identifies key areas for improvement. Future research can focus on enhancing models' ability to more accurately capture visual elements and interpret figurative language. 

\section{Ethical Considerations and Limitations}

\paragraph{Content Hallucinations and Inaccuracies.} One critical concern is that the model might generate irrelevant or inaccurate interpretations of memes \cite{maynez2020faithfulness, ji2023survey}, which could inadvertently perpetuate stereotypes or biases about certain social groups. This issue is inherent in the use of LMMs in a zero-shot manner, where the model operates without specific training on the task at hand. In our work, we address this challenge by focusing on enhancing explainability behind model decisions, aiming to provide more transparent reasoning for the outputs generated. However, the limitations associated with hallucinations highlight the need for future research to explore more robust approaches, such as retrieval-augmented generation, which could improve the accuracy and relevance of generated interpretations. This would not only enhance the model's performance but also mitigate potential ethical risks associated with the propagation of harmful stereotypes.

\paragraph{Generalisability to New Unseen Memes.} When deploying fine-tuned models for hateful meme detection, a primary ethical concern is their ability to generalize effectively to unseen memes, which can lead to the transfer of domain-specific biases and subsequent misclassification \cite{cao2024modularized}. To address this challenge, our framework employs LMMs to generate meme interpretations in a zero-shot manner. By avoiding fine-tuning for specific domains, these LMMs are less prone to overfitting and perpetuating biases against particular social groups. This approach allows our framework to leverage the strengths of generalized LMMs while minimizing the risk of bias. However, we acknowledge that these generalized models may still harbor inherent biases, presenting ethical risks in the context of automated hateful meme detection. Therefore, ongoing vigilance and evaluation are necessary to ensure that our framework operates equitably and responsibly in real-world applications.

\paragraph{Misuse of Meme Interpretations.} While these interpretations are designed to enhance understanding and assist in content moderation, we acknowledge the risk that they could be misused to create more hateful memes and reinforce social stereotypes. We strongly condemn such actions and want to clarify that we intend to use these interpretations to improve content moderation. We believe that the benefits of generating meme interpretations for this purpose far outweigh any potential risks of misuse. By providing content moderators with deeper insights, we aim to empower them to identify and flag potentially hateful content more effectively, thereby contributing to a more informed and responsible digital environment.


\bibliography{aaai22}

\subsection{Paper Checklist to be included in your paper}

\begin{enumerate}

\item For most authors...
\begin{enumerate}
    \item  Would answering this research question advance science without violating social contracts, such as violating privacy norms, perpetuating unfair profiling, exacerbating the socio-economic divide, or implying disrespect to societies or cultures?
    \answerYes{Yes, our work primarily focuses on utilizing LMMs to analyze and generate interpretations of hateful memes. While these generated interpretations may reflect social stereotypes, our goal is to enhance hateful meme detection systems and improve the understanding of such content.}
  \item Do your main claims in the abstract and introduction accurately reflect the paper's contributions and scope?
    \answerYes{Yes.}
   \item Do you clarify how the proposed methodological approach is appropriate for the claims made? 
    \answerYes{Yes.}
   \item Do you clarify what are possible artifacts in the data used, given population-specific distributions?
    \answerYes{Yes.}
  \item Did you describe the limitations of your work?
    \answerYes{Yes. You may find them under "Ethical Considerations and Limitations" section}
  \item Did you discuss any potential negative societal impacts of your work?
    \answerYes{Yes. You may find them under "Ethical Considerations and Limitations" section}
      \item Did you discuss any potential misuse of your work?
    \answerYes{Yes. You may find them under "Ethical Considerations and Limitations" section}
    \item Did you describe steps taken to prevent or mitigate potential negative outcomes of the research, such as data and model documentation, data anonymization, responsible release, access control, and the reproducibility of findings?
    \answerNA{N/A}
  \item Have you read the ethics review guidelines and ensured that your paper conforms to them?
    \answerYes{Yes.}
\end{enumerate}

\item Additionally, if your study involves hypotheses testing...
\begin{enumerate}
  \item Did you clearly state the assumptions underlying all theoretical results?
    \answerNA{N/A}
  \item Have you provided justifications for all theoretical results?
    \answerNA{N/A}
  \item Did you discuss competing hypotheses or theories that might challenge or complement your theoretical results?
    \answerNA{N/A}
  \item Have you considered alternative mechanisms or explanations that might account for the same outcomes observed in your study?
    \answerNA{N/A}
  \item Did you address potential biases or limitations in your theoretical framework?
    \answerNA{N/A}
  \item Have you related your theoretical results to the existing literature in social science?
    \answerNA{N/A}
  \item Did you discuss the implications of your theoretical results for policy, practice, or further research in the social science domain?
    \answerNA{N/A}
\end{enumerate}

\item Additionally, if you are including theoretical proofs...
\begin{enumerate}
  \item Did you state the full set of assumptions of all theoretical results?
    \answerNA{N/A}
	\item Did you include complete proofs of all theoretical results?
    \answerNA{N/A}
\end{enumerate}

\item Additionally, if you ran machine learning experiments...
\begin{enumerate}
  \item Did you include the code, data, and instructions needed to reproduce the main experimental results (either in the supplemental material or as a URL)?
    \answerYes{The GitHub link can be found in the paper's abstract.}
  \item Did you specify all the training details (e.g., data splits, hyperparameters, how they were chosen)?
    \answerYes{Yes. These information can be found under "Implementation Details" section.}
     \item Did you report error bars (e.g., with respect to the random seed after running experiments multiple times)?
    \answerYes{Yes. These details can be found in Table 3 and 4, where the model performance over multiple seeds have been reported.}
	\item Did you include the total amount of compute and the type of resources used (e.g., type of GPUs, internal cluster, or cloud provider)?
    \answerYes{Yes. These information can be found under "Implementation Details" section.}
     \item Do you justify how the proposed evaluation is sufficient and appropriate to the claims made? 
    \answerYes{Yes. These information can be found under "Experiment Results" section.}
     \item Do you discuss what is ``the cost`` of misclassification and fault (in)tolerance?
    \answerNA{N/A}
  
\end{enumerate}

\item Additionally, if you are using existing assets (e.g., code, data, models) or curating/releasing new assets, \textbf{without compromising anonymity}...
\begin{enumerate}
  \item If your work uses existing assets, did you cite the creators?
    \answerYes{Yes.}
  \item Did you mention the license of the assets?
    \answerNA{N/A.}
  \item Did you include any new assets in the supplemental material or as a URL?
    \answerNA{N/A.}
  \item Did you discuss whether and how consent was obtained from people whose data you're using/curating?
    \answerNA{N/A.}
  \item Did you discuss whether the data you are using/curating contains personally identifiable information or offensive content?
    \answerNA{N/A.}
\item If you are curating or releasing new datasets, did you discuss how you intend to make your datasets FAIR?
\answerNA{N/A.}
\item If you are curating or releasing new datasets, did you create a Datasheet for the Dataset? 
\answerNA{N/A.}
\end{enumerate}

\item Additionally, if you used crowdsourcing or conducted research with human subjects, \textbf{without compromising anonymity}...
\begin{enumerate}
  \item Did you include the full text of instructions given to participants and screenshots?
    \answerNA{N/A.}
  \item Did you describe any potential participant risks, with mentions of Institutional Review Board (IRB) approvals?
    \answerNA{N/A.}
  \item Did you include the estimated hourly wage paid to participants and the total amount spent on participant compensation?
    \answerNA{N/A.}
   \item Did you discuss how data is stored, shared, and deidentified?
   \answerNA{N/A.}
\end{enumerate}

\end{enumerate}

\end{document}